\documentclass{article}

\PassOptionsToPackage{numbers,square}{natbib}
\usepackage[final]{neurips_2022_ml4ad}

\usepackage[utf8]{inputenc} 
\usepackage[T1]{fontenc}    
\usepackage[backref]{hyperref}
\usepackage{url}            
\usepackage{booktabs}       
\usepackage{multirow}
\usepackage{amsfonts}       
\usepackage{nicefrac}       
\usepackage{microtype}      
\usepackage{xcolor}         
\usepackage{soul}
\usepackage[pdftex]{graphicx} 
\usepackage{wrapfig}        
\usepackage{amsmath}        
\usepackage{subcaption}
\usepackage{float}
\DeclareMathOperator*{\argmin}{arg\,min}
\title{Controlling Steering with Energy-Based Models}

\author{%
  Mikita Balesni \\
  Autonomous Driving Lab \\
  University of Tartu \\
  \texttt{mbalesni@gmail.com} \\
  \And
  Ardi Tampuu \\
  Autonomous Driving Lab \\
  University of Tartu \\
  \texttt{ardi.tampuu@ut.ee} \\
  \AND
  Tambet Matiisen \\
  Autonomous Driving Lab \\
  University of Tartu \\
  \texttt{tambet.matiisen@ut.ee} \\
}

\begin{document}

\maketitle

\begin{abstract}
  So-called implicit behavioral cloning with energy-based models has shown promising results in robotic manipulation tasks. We tested if the method's advantages carry on to controlling the steering of a real self-driving car with an end-to-end driving model. We performed an extensive comparison of the implicit behavioral cloning approach with explicit baseline approaches, all sharing the same neural network backbone architecture. Baseline explicit models were trained with regression (MAE) loss, classification loss (softmax and cross-entropy on a discretization), or as mixture density networks (MDN). While models using the energy-based formulation performed comparably to baseline approaches in terms of safety driver interventions, they had a higher whiteness measure, indicating higher jerk. To alleviate this, we show two methods that can be used to improve the smoothness of steering. We confirmed that energy-based models handle multimodalities slightly better than simple regression, but this did not translate to significantly better driving ability. We argue that the steering-only road-following task has too few multimodalities to benefit from energy-based models. This shows that applying implicit behavioral cloning to real-world tasks can be challenging, and further investigation is needed to bring out the theoretical advantages of energy-based models.
\end{abstract}

\section{Introduction}

\begin{figure}[t]
    \centering
    \includegraphics[width=0.9\linewidth]{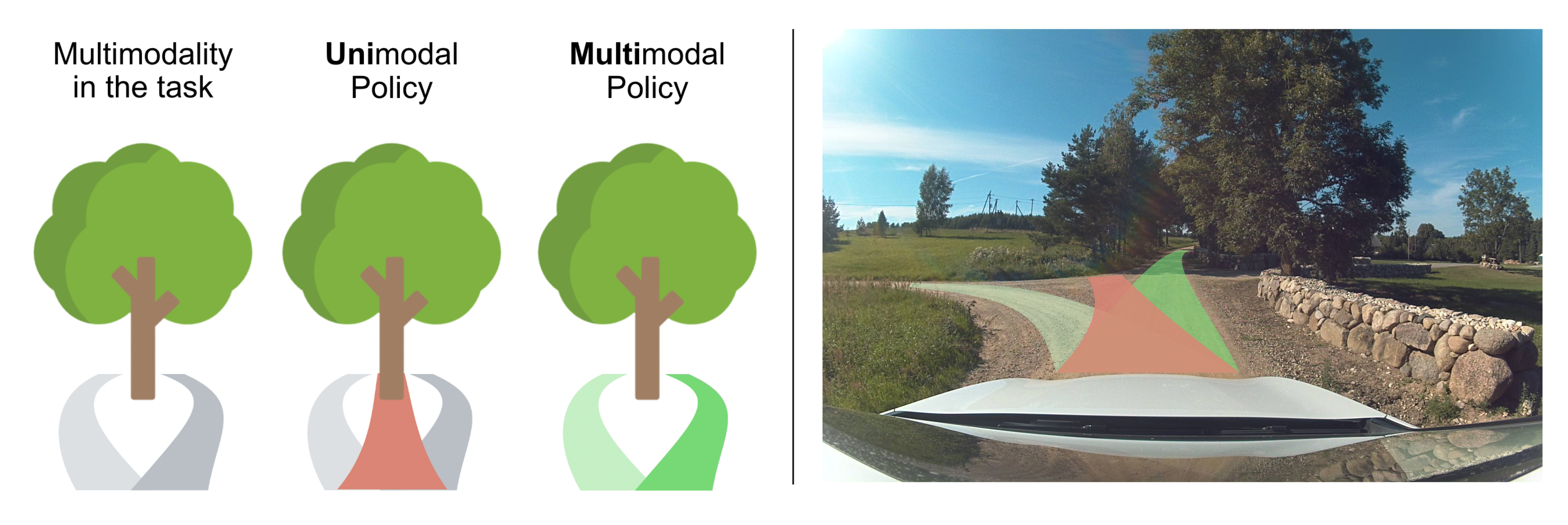}
    \caption{Left: If the experts have passed the tree from left and right with equal frequency in the training data, behavioral cloning with a unimodal policy and no high-level navigation commands would average the training trajectories and drive straight into the tree. Right: In practice, we have experienced such behavior at locations where side roads enter the main road - unimodal regression models tend to swerve slightly (the red trajectory) towards the side road. These swerves are minor because side roads are rarely taken in our training data, and keeping straight is the dominant behavior. }
    \label{fig:multimodality-illustration}
\end{figure}

Implicit behavioral cloning\cite{florence2021implicit} with energy-based models \cite{lecun-ebm} has shown a lot of promise in robotic manipulation tasks. The theoretical advantages of energy-based models include increased data efficiency and the ability to model discontinuities and multimodalities in the output action distribution \cite{florence2021implicit}. Here, we set out to evaluate energy-based models for controlling the steering of a self-driving vehicle using end-to-end driving models \cite{tampuu2020survey,bojarski2016end}. We work with steering-only models as the usefulness of multimodality-handling is evident in this output modality, as illustrated by the theoretical and practical failure cases of unimodal models on Figure \ref{fig:multimodality-illustration}. Adding longitudinal control would complicate the task and demand more training data while not necessarily adding more multimodal situations.

To validate the theoretical advantages empirically, we compare a simple energy-based model (EBM) with several baseline approaches in a road-following task in the real world. The main experiments were also repeated in the VISTA\cite{amini2022vista} simulator. The explicit baseline models are based on the same neural network architecture as the EBM, with only the necessary modifications. They are trained with the MAE loss, classification loss (softmax with cross-entropy), or as mixture density networks \cite{bishop1994mixture} (MDN). During on-policy testing, the models controlled only the car's steering; the location- and direction-relevant velocity was taken from a previously recorded expert trajectory. The evaluation was performed on a WRC 2022 Rally Estonia track designed to be challenging for humans, which was not included in the training set. 

According to the main evaluation metric, safety-driver intervention count, energy-based models performed comparably to the baseline methods but had noticeably jerkier steering. To alleviate this, we proposed two methods: temporal smoothing of predicted steering angles and spatially-aware soft targets for cross-entropy loss. Still, unimodal explicit behavioral cloning performed best in terms of safety driver interventions and jerk. We argue that the inductive bias enforced by unimodal losses makes them more data-efficient in simpler road situations, but more data is needed to model situations requiring true multimodalities.

The main contributions of the paper are as follows:

\textbf{1.} We show that controlling the steering of an autonomous vehicle with energy-based models in the real world performs comparably to the baseline explicit behavioral cloning approaches.

\textbf{2.} We propose two methods that effectively reduce the steering jerk of energy-based models: temporal smoothing of predicted steering angles and soft targets for the cross-entropy loss.

\textbf{3.} To our surprise, we find that energy-based models do not outperform any of the similar-architecture explicit behavioral cloning baseline approaches in the real-world road-following task and that representing multimodalities does not translate into better driving.

\section{Background}

\paragraph{End-to-end driving}
End-to-end driving attempts to replace the classical modular self-driving pipeline with a single neural network model \cite{tampuu2020survey}. In the purest form, an end-to-end self-driving model takes in raw sensor data and yields actionable commands such as steering angle, throttle, and brake values. Such models are commonly trained with behavioral cloning \cite{pomerleau1989alvinn,bojarski2016end} to imitate human expert commands in the same situation.

One of the popular end-to-end driving models is NVIDIA PilotNet \cite{bojarski2016end}. In our work, we use the PilotNet network architecture as the backbone of all models because it is relatively fast to train and sufficient for the road-following task we aimed for. We do not use conditioning on high-level commands \cite{codevilla2018end} or more complex network architectures \cite{codevilla2019exploring} to keep the setup simple - our goal is to compare similar-capacity implicit and explicit models, not to aim for the best performance. We use steering angle as the network output, rather than trajectory\cite{bojarski2020nvidia,hawke2020urban} or costmap\cite{zeng2019end}, to be in line with prior implicit learning work in robotics \cite{florence2021implicit} where the network predicted raw actuator signal.

\begin{figure}[t]
    \centering
    \includegraphics[width=0.8\linewidth]{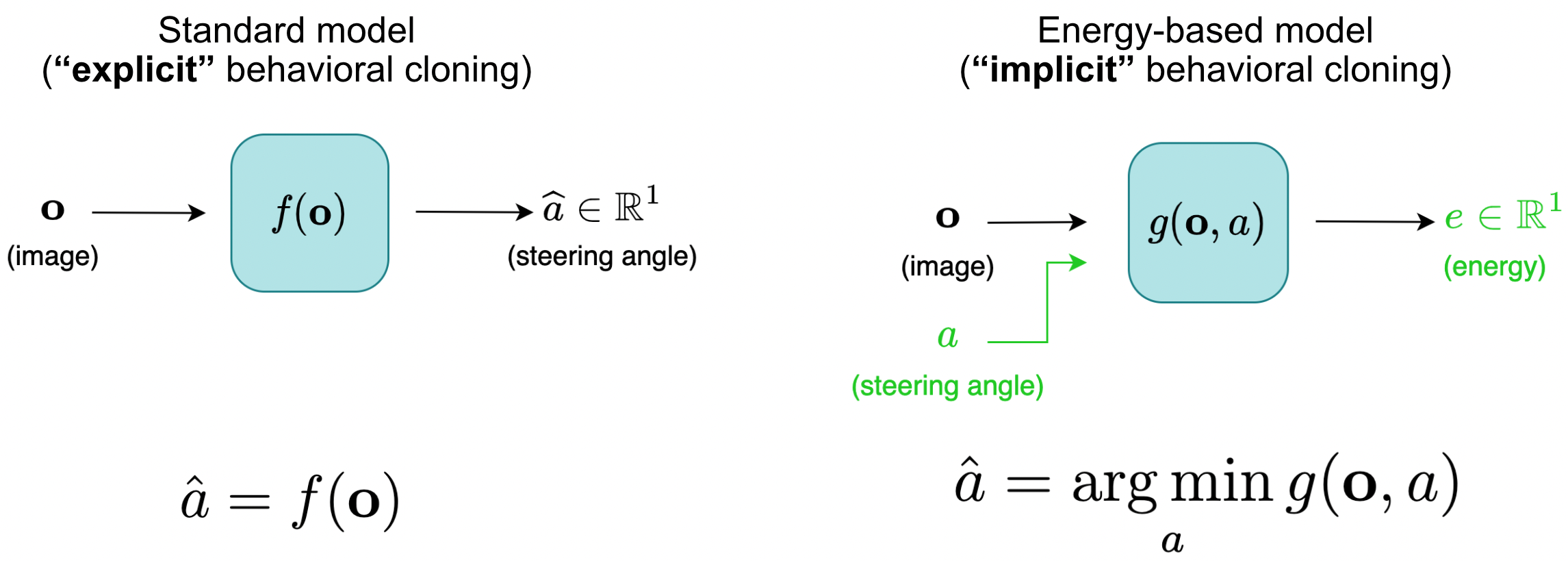}
    \caption{Controlling steering with implicit behavioral cloning. Left: explicit models output the predicted angle directly. Right: implicit energy-based models return an energy value per observation-action pair. The action yielding the lowest energy for the current observation is chosen via \textit{argmin}.}
    \label{fig:implicit-vs-explicit}
\end{figure}

A significant amount of end-to-end driving research is done in simulations\cite{dosovitskiy2017carla}, which through repeatability and access to the state of the world, allow benchmarking with more complex measures of driving quality \cite{codevilla2019exploring,carlaLeaderboard}. However, models created in simulations cannot be deployed in the real world without adaptations \cite{bewley2019learning}. Also, some problems related to steering delays and passenger comfort only become apparent in the real world. Here, we have sufficient data for learning the relatively simple road-following task, so we choose to work in the real world. 

While the data-driven approach to autonomous driving is viewed as the most promising path to full autonomy by some authors \cite{jain2021autonomy, hawke2021reimagining}, significant problems remain to be solved, most prominently in generalization, in the explainability of decisions, and in providing safety guarantees \cite{codevilla2019exploring,nassi2020phantom,kalra2016driving}.

\paragraph{Energy-based models}
An  \textit{energy function}, described by \citet{lecun-ebm}, is any continuous function that measures "goodness" between two sets of variables, where "good" pairs have a low energy value. Following \citet{florence2021implicit} who coined the term, we call behavioral cloning policies "implicit" when they are composed of \textit{argmin} and a continuous energy function $E$, such that:

\begin{equation}
    \mathbf{\hat{a}}=\argmin_{\mathbf{a} \in \mathcal{A}} E(\mathbf{o},\mathbf{a}),
\end{equation}

where $\mathbf{o}$ is an observation, e.g., a camera image, and $a$ is an action, e.g., a steering wheel angle. In the present work, $E$ is implemented by a neural network with PilotNet architecture with minor modifications (discussed below). The classical approach of a model directly computing the action based on an observation is in this context called "explicit" behavioral cloning (Figure \ref{fig:implicit-vs-explicit}).

Implicit behavioral cloning with energy-based models promises the following three advantages compared to classical explicit behavioral cloning: the ability to represent discontinuities sharply, the ability to represent multimodal action distributions, and better generalization with improved data efficiency. In this work, we mainly focus on the ability to model multimodalities.

\paragraph{Evaluation of driving models} 

With behavioral cloning, models are optimized to make momentary decisions on data originating from the distribution resulting from human driving. When deployed, however, the solutions face a sequential decision-making task on data originating from a distribution caused by their own driving. Off-policy metrics computed on held-out datasets of expert driving, such as mean absolute error (MAE), measure only the predictive ability of the models. However, such measures are insufficient for predicting success at the sequential decision-making task when deployed \cite{codevilla2018offline}. Despite modest correlations with driving ability, we use these metrics for model selection, as is often done in related works.

Among the on-policy metrics measured during model deployment, the number of safety driver interventions, the mean distance between interventions (DBI), and the amount of time or distance traveled autonomously are the most popular metrics \cite{tampuu2020survey}. In our work, we chose the number of interventions as the main metric, as the distance traveled was fixed.

Beyond just completing routes safely, the comfort of passengers matters. The smoothness of driving has been related directly to passenger comfort and perceived safety\cite{hecker2020learning,elbanhawi2015passenger}. We follow multiple previous works \cite{https://doi.org/10.48550/arxiv.1710.03804,https://doi.org/10.48550/arxiv.1811.05785,lidar-as-camera} that quantify the smoothness of steering with \textit{whiteness}, defined as:

\begin{equation}
    W = \sqrt{\frac{1}{D} \sum_{i=1}^{D} \Big(\frac{\delta P_i}{\delta t}\Big)^2},
\end{equation}

where $\delta P_{i}$ is the steering angle change, $D$ is dataset size, and $\delta t$ is the time between decisions.


\section{Methods}

\begin{figure}[t]
    \centering
    \includegraphics[width=0.9\linewidth]{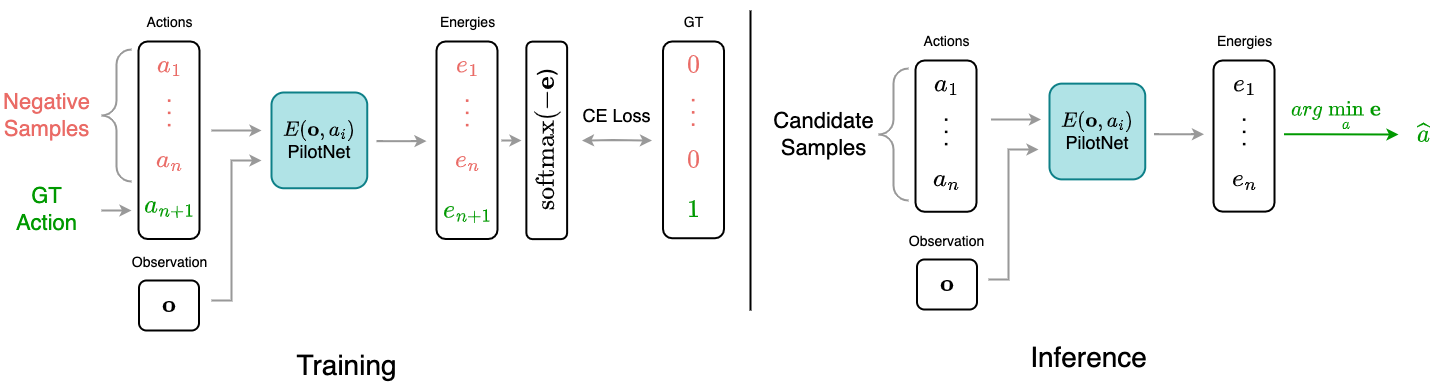}
    \caption{Energy-based model training and inference procedures. Left: feeding the observation and different action values (one action at a time) through the energy-computing network results in a vector of energy values, which is then optimized via CE loss to make the ground truth action have the lowest energy. Right: Observation is fed through the network with different candidate action values, producing an energy value per candidate; the lowest-energy action is chosen.}
    \label{fig:ebm-trainin-inference}
\end{figure}

\paragraph{Baseline models}
We adapt the PilotNet architecture's output layer to produce a variety of baseline models. The regression baseline has just one output node optimized to produce the steering value via MAE loss. The classification baseline has the action space divided into N bins and optimizes predicting the right bin via cross-entropy loss. Mixture density networks output between 1 to 5 triplets of mean, standard deviation, and relevancy scores $\alpha_i$ ($\alpha_i \geq 0, \sum_i \alpha_i = 1$), a linear combination of which produces a Gaussian mixture model over action values. We use the mean of the most likely Gaussian during deployment.

\paragraph{EBM Training and Inference}

We adapt the training and inference algorithms from energy-based model literature \cite{florence2021implicit} with a few modifications (see Figure \ref{fig:ebm-trainin-inference}). As a first modification, we use a constant grid of linearly-spaced steering angles durig training and inference instead of sampling uniformly. Second, we do not use inference-time optimization to improve the initial candidate actions. The candidate action values cover the steering angle range densely, and further optimization yielded no gains. Furthermore, a fixed set of values is required by one of the proposed EBM modifications and helps to make a cleaner comparison with the classification baseline. Early offline experiments (see Appendix Figure \ref{fig:ebm-variants-performance}) showed that these changes resulted in at least as good performance on steering prediction as random sampling and inference-time optimization.

This results in the following loss function:
\begin{equation}
L_{EBM} = \text{CE}(\text{softmax}(\mathbf{-e}),\mathbf{y})= - \sum_{i=1}^{n+1}(y_i \cdot log(-\frac{e^{e_i}}{\sum_{j=1}^{n+1} e^{e_j}})), 
\end{equation}

where $\mathbf{e}$ is the vector of energy values produced by neural network outputs $e_i = E(\mathbf{o},a_i)$ and $\mathbf{y}$ is one-hot vector having 1 at the position of ground truth action. Both $\mathbf{e}$ and $\mathbf{y}$ contain $n+1$ elements: $n$ sampled values and one ground truth. To get the action energy vector $\mathbf{e}$ for a single observation, the neural network $E$ is run on a batch of samples with the same repeated observation $\mathbf{o}$ and different actions $a_i$. In practice, the convolutional part of the network runs on an image only once, with actions fused into the model before the MLP head to reduce the time and memory usage.

Our initial implementation of EBM demonstrated a high whiteness score, i.e., high lateral jerk. This characteristic did not depend on the amount of training data (see Appendix Figure \ref{fig:mae-whiteness-scaling}). We explored two changes to the EBM to combat this undesired characteristic.

\paragraph{EBM with Temporal Smoothing}

If one could reduce a model's sensitivity to slight differences in subsequent camera frames, one would achieve temporally smoother predictions. An obvious choice in the case of energy-based models is to minimize the difference between predicted energy distributions at subsequent frames. So, we propose adding a temporal smoothness loss term, defined as:

\begin{equation}
    L_{temp} = \alpha \Big{\Vert}\mathbf{e}_t - \mathbf{e}_{t+1}\Big{\Vert},
\end{equation}

where $\mathbf{e}_t$ stands for the vector of predicted energy values at timestep $t$, and $\alpha$ is the smoothing strength. EBMs take actions as input, so  $\mathbf{e}_t$ and $\mathbf{e}_{t+1}$ have to be computed with the same steering angle inputs, which motivated our use of a constant action grid instead of random sampling. However, to stick with the conventional sampling, one could also draw a random sample \textit{once per pair} of frames. Since the ground truth steering angle is often different for consecutive frames, its energy values are masked in this loss term. A range of well-performing smoothing strengths was found empirically. We use $\alpha=1.0$ for the temporally smoothed EBM in the final experiments.

\paragraph{EBM with Soft Targets}
We hypothesized that using one-hot targets in the cross-entropy loss is a major contribution to the higher whiteness of EBMs. Forcing nearby steering values to have drastically different energy is likely to make learning less efficient and the energy landscape noisier. This can lead to higher variance when choosing the best action via \emph{argmin}.

\begin{figure}[h]
    \centering
    \includegraphics[width=0.95\linewidth]{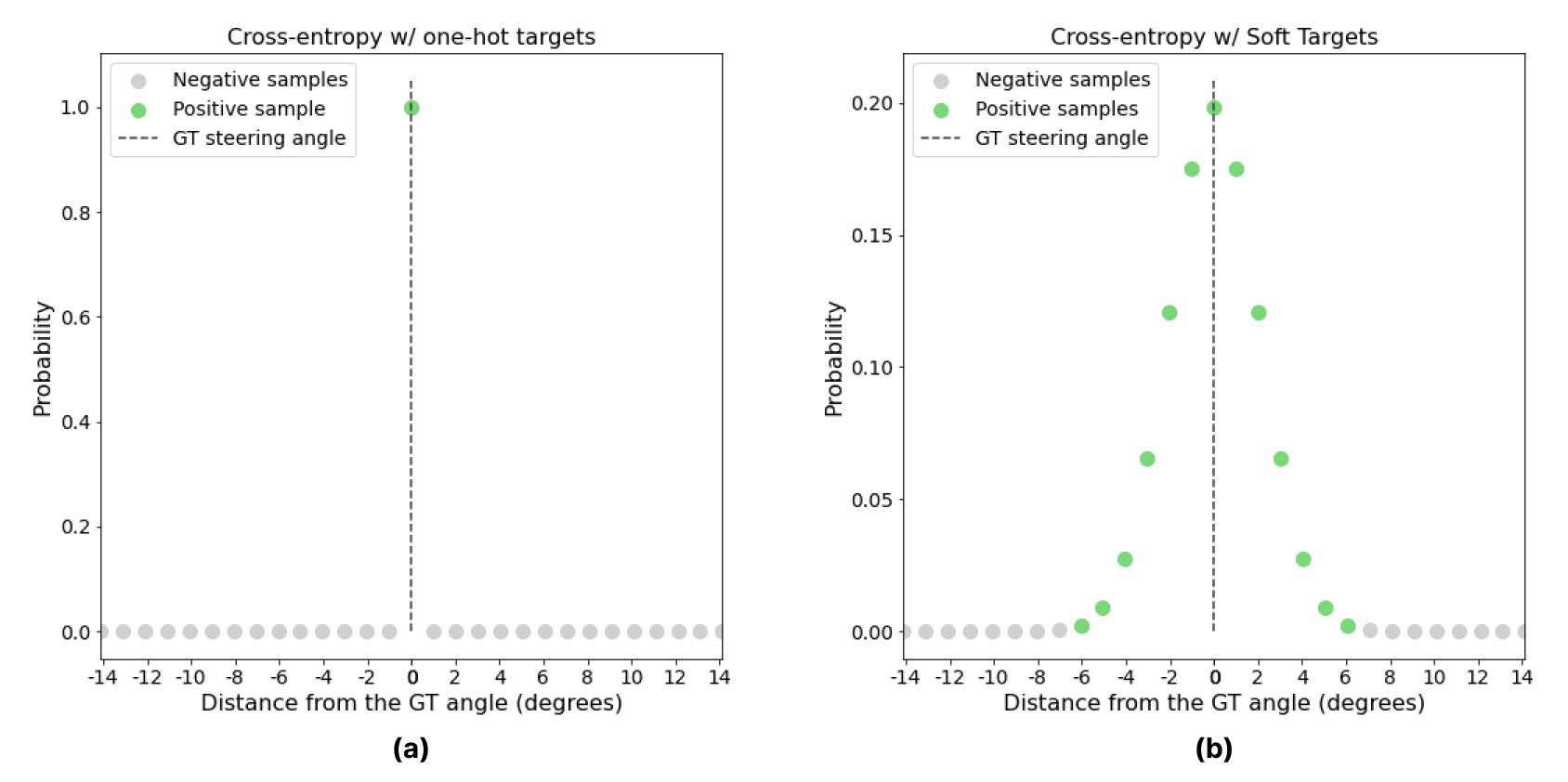}
    \caption{\textbf{(a)} One-hot (standard) cross-entropy pushes the probability of negative samples down, even if they are indistinguishably close to the target. \textbf{(b)} Soft targets give positive weights to the samples around the target proportionally to the distance. Note that the horizontal axis is trimmed to focus on the GT value; the full range of input actions we use is from $\sim$-250 to $\sim$250 degrees.}
    \label{fig:soft-vs-one-hot-targets}
\end{figure}

Hence, we investigated a simple fix: use soft targets for the cross-entropy loss. Whereas soft targets have been widely used in neural networks with the purpose of regularization and better calibration \cite{pereyra2017regularizing,muller2019does}, our use case is a bit different.  Unlike usual classification targets, our outputs are ordinal, and we aim to enforce spatial smoothness. We replace the one-hot ground-truth vector with a vector assigning some of the probability to actions a few degrees away from the ground truth (see Figure \ref{fig:soft-vs-one-hot-targets}). Target probabilities are computed as:

\begin{equation}
    \mathbf{p^*} = \text{softmax}\Big(\frac{-(\mathbf{a}-a_{GT})^2}{T}\Big),
\end{equation}

where $\mathbf{p^*}$ is a vector of target probabilities for cross-entropy loss, $\mathbf{a}$ is the vector of input candidate steering values, $a_{GT}$ is the ground truth steering angle, and $T$ is the softmax temperature ($2.5 * 10^{-3}$ in all reported tests with soft targets). We picked the temperature value such that $99.9\%$ of the probability mass was on $\pm5$ degrees around the ground truth.

\section{Experimental Setup}

\paragraph{Dataset and Training Pipeline}

We use the training dataset by \citet{lidar-as-camera}, which consists of 540 km of human driving on WRC Rally Estonia tracks. These are usually very low-traffic gravel roads. The recordings that make up the dataset are broken into training (460 km) and evaluation (80 km), with the evaluation recordings used for off-policy metric calculation and early stopping. The original dataset includes camera and LiDAR images, but only camera frames are used in our experiments. Image pre-processing and training details are specified in the supplementary materials.

\paragraph{On-Policy Evaluation}

The on-policy evaluation was performed on a 4.3 km section of the WRC Estonia 2022 SS10+14 Elva track \footnote{https://www.rally-maps.com/Rally-Estonia-2022/Elva}, driven in both directions. No recordings from this track were in the training set of the models. The speed was set to 80\% of the speed a human driver used in the same location and direction, extracted from a prior recording. In practice, this meant a speed of up to 40 km/h. Setting the speed to 100\% was attempted but felt too dangerous with certain models. The testing was performed over multiple weeks of September 2022.

The evaluation track is narrow, and driving off the road edge is hazardous for the car, so the safety driver was free to intervene when they perceived danger. An intervention was counted when the driver applied force to turn the steering wheel. If the model turned the steering wheel simultaneously in the same direction as the safety driver, it would not cause an intervention since no force was applied.

Alongside the intervention count, we also report the whiteness of steering as an on-policy metric. Here, \emph{command whiteness} $W_{cmd}$ stands for the whiteness of the predicted steering commands during on-policy evaluation, and  \emph{effective whiteness} $W_{eff}$ refers to the resulting actual whiteness of the front wheel angles as measured by the sensors. Command whiteness is usually higher than effective whiteness due to the smoothing effect of the real-car actuators. No matter the force (i.e., the angular acceleration of the steering wheel), it takes time to reach the target value.

First, around twenty test runs were completed to select the best representative of each model type over several hyperparameters and random seeds. For the final experiments, the six most promising models were chosen: an EBM with 512 linearly-spaced candidate values (standard, with temporal smoothing, or with soft targets), a classification model with 512 bins, MDN with 5 Gaussians, and a regression model with MAE loss. We performed four evaluation runs per model across four days. The worst run for each model was discarded to account for out-of-distribution weather (excessive sun or rain drops on camera) or safety driver variance.

\paragraph{Evaluation in VISTA}

Evaluating driving models in the real world can make it harder for other researchers to replicate results. To aid reproducibility, we additionally run the main experiments in the VISTA Driving Simulator\cite{amini2022vista}. VISTA is a data-driven simulator that allows replaying recordings of real-world drives \textit{interactively} by reprojecting the viewpoint as desired. Thus, a simulator can be used for on-policy, closed-loop evaluation, allowing fast and reproducible model evaluation (as in standard model-based simulators) while staying visually close to the real-world data distribution. To this end, we release our evaluation code and the recording we used for evaluation in the VISTA format.\footnote{https://github.com/UT-ADL/vista-evaluation} We used a recording produced by the strongest model completing the track without interventions. This recording had the highest correlation with our results on all models (see Appendix Table \ref{table:vista-correlations}), due to being most in-distribution with the weather and vegetation at the time of the real-world tests.

Absent a safety driver, we define crashes in VISTA as moments when the car drives more than 2 meters away from the expert-driven trajectory. After a crash, we restart the car two seconds further down the road. This evaluation scheme has obvious limitations, for example, as the expert does not always drive in the center of the road, and 2 meters would be too late or too soon to cause the safety driver to disengage in reality. Yet, empirical results from the evaluation in VISTA support the findings from our real-world experiments, suggesting that VISTA can be used to reproduce our key results.

\section{Results} \label{results-section}

\begin{table}[h]
\caption{Generalization results, with three real-world and three virtual driving sessions per model.}
\label{generalization-results-table}
\centering
\begin{tabular}{llllll}
\toprule
          &  \multicolumn{3}{c}{Real world} & \multicolumn{2}{c}{VISTA} \\
          \cmidrule(r){2-4} \cmidrule(r){5-6}
          Model &  Interventions & $W_{eff}$ & $W_{cmd}$ & Crashes & $W_{cmd}$ \\
\midrule
          \multirow{3}{*}{EBM} & 4  & 35.25°/s & 176.93°/s & 2    & 114.33°/s \\
           &                      1 & 32.34°/s & 96.94°/s  & 1    & 121.57°/s \\
           &                      2 & 28.57°/s & 223.59°/s & 2    & 121.67°/s \\
          \cmidrule(r){2-4} \cmidrule(r){5-6}
          \qquad\qquad\qquad mean: &  2.33 & 32.05°/s & 165.82°/s & 1.67 & 119.19°/s \\

          \midrule
          \multirow{3}{*}{EBM Temp. Smoothing} &   5  & 49.92°/s & 119.39°/s & 3 & 58.70°/s \\
           &                                       2  & 38.96°/s & 137.22°/s & 2 & 60.37°/s \\
           &                                       3  & 34.21°/s &  77.28°/s & 2 & 48.86°/s  \\
           \cmidrule(r){2-4} \cmidrule(r){5-6}
           \qquad\qquad\qquad mean:                &   3.33  & 41.03°/s & 111.30°/s & 2.33 & 55.98°/s \\

           \midrule
           \multirow{3}{*}{EBM Soft Targets}     &  5 & 27.80°/s & 56.33°/s & 3 & 85.72°/s \\
           &                                        5 & 46.83°/s & 57.15°/s & 3 & 74.97°/s \\
           &                                        4 & 33.72°/s & 56.86°/s & 3 & 81.87°/s \\
           \cmidrule(r){2-4} \cmidrule(r){5-6}
           \qquad\qquad\qquad mean:                  &  4.66 & 36.12°/s & 56.78°/s & 3 & 80.85°/s \\

           \midrule
           \multirow{3}{*}{Regression (MAE)} &      2 & 26.69°/s & 37.84°/s & 0 & 24.39°/s \\
           &                                        2 & 29.65°/s & 75.34°/s & 0 & 24.75°/s \\
           &                                        1 & 26.28°/s & 33.10°/s & 0 & 24.25°/s \\
           \cmidrule(r){2-4} \cmidrule(r){5-6}
           \qquad\qquad\qquad mean:                  &  1.66 & 27.54°/s & 48.76°/s & 0 & 24.47°/s \\

           \midrule
           \multirow{3}{*}{Classification} &        1 & 41.05°/s & 182.39°/s & 1 & 123.69°/s \\
           &                                        7 & 62.17°/s & 287.14°/s & 1 & 105.13°/s \\
           &                                        1 & 34.11°/s & 162.27°/s & 1 & 104.31°/s \\
           \cmidrule(r){2-4} \cmidrule(r){5-6}
           \qquad\qquad\qquad mean:                   & 3.00 & 45.77°/s & 210.60°/s & 1 & 111.04°/s \\
         
           \midrule
           \multirow{3}{*}{MDN} &                   1 & 25.32°/s & 33.62°/s & 3 & 37.22°/s \\
           &                                        5 & 24.82°/s & 35.46°/s & 3 & 35.74°/s \\
           &                                        5 & 26.66°/s & 37.39°/s & 3 & 35.84°/s \\
           \cmidrule(r){2-4} \cmidrule(r){5-6}
           \qquad\qquad\qquad mean:                   & 3.66 & 25.59°/s & 35.49°/s & 3 & 36.27°/s \\
\bottomrule
\end{tabular}
\end{table}

The results of the final test runs are presented in Table \ref{generalization-results-table}. Each row in the table corresponds to approximately 20 minutes of driving, so each model's total intervention count is produced by an hour of driving. We believe this is enough time to reveal noteworthy differences in performance on a simple task such as road following. Real-world results are supported by VISTA showing similar trends with lower variance (Pearson R = 87.5\%, Spearman R = 92.7\% for real-world interventions in final experiments vs VISTA crashes).

We observe no clear benefit of using energy-based implicit behavioral cloning models over the explicit baseline models. The baseline regression model resulted in the least interventions and had also the most smooth driving according to the whiteness measure. The other two explicit baseline models: classification and mixture density networks, resulted in more interventions across the three runs than the EBM. Supported by VISTA, the results reveal that no one solution stands out clearly, and hence that using the EBM formulation did not significantly improve the performance.

To bring the whiteness values of EBMs closer to the values of explicit models, we implemented two variations to the EBM. Both approaches significantly reduced the whiteness of the model predictions during deployment ($W_{cmd}$ in Table \ref{generalization-results-table}). However, this reduction did not translate into a reduction of effective whiteness, while both of these approaches resulted in more safety driver interventions.

\begin{figure}[h]
    \centering
    \includegraphics[width=1\linewidth]{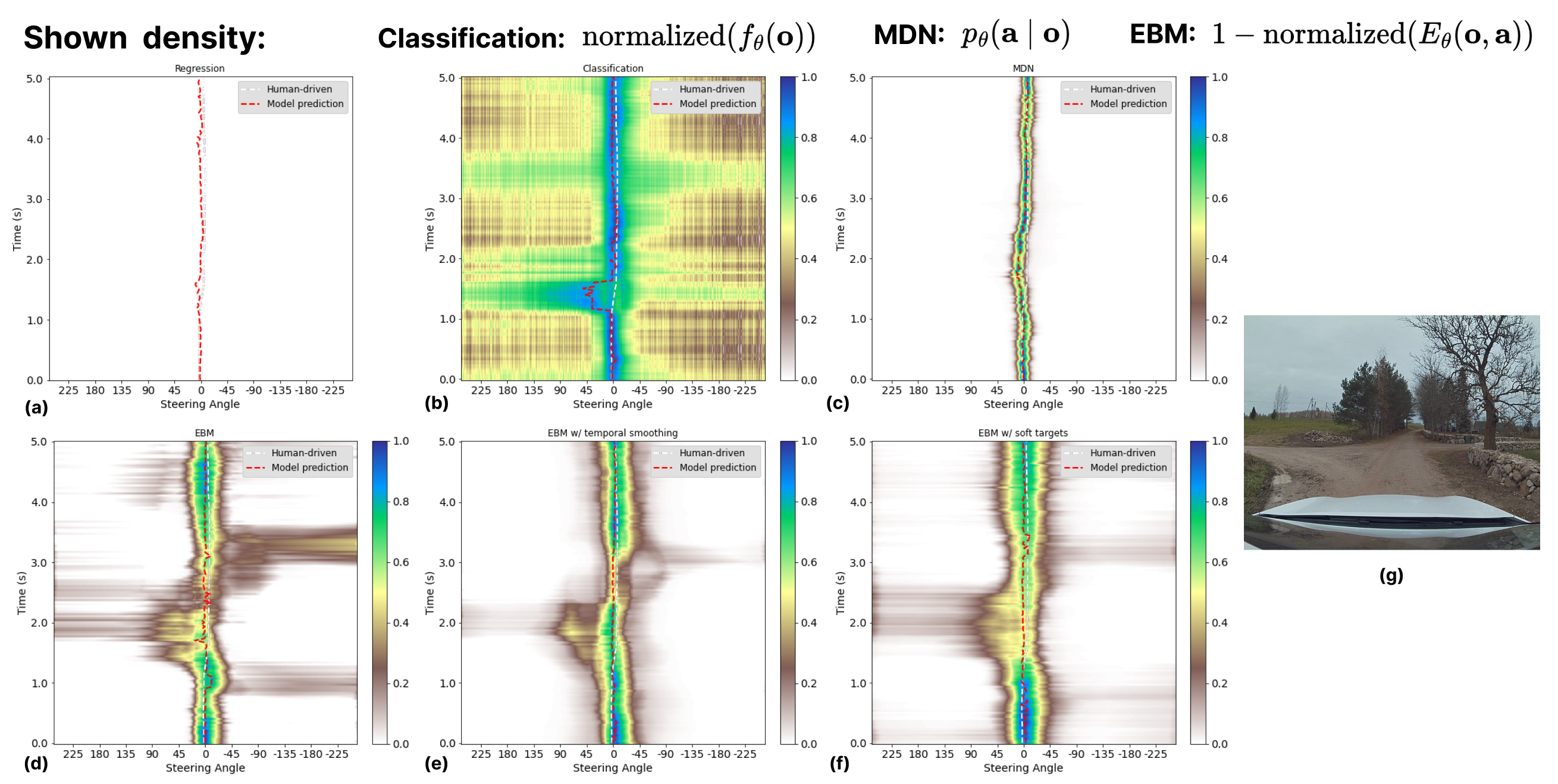}
    \vspace{-0.2cm}
    \caption{\textbf{Outputs of different models during 5 seconds of human driving (a-f).} The period corresponds to the car passing an intersection, with Y-axis representing the time and X-axis the predicted steering angle. The camera image 2 seconds into the recording is given on the right (g). The red dashed line represents the predicted steering values, grey dashed line is the ground truth.}
    \label{fig:intersection-density-plots}
\vspace{-0.2cm}
\end{figure}

The increased temporal stability of modified EBM approaches compared to the naive EBM is also visible in Figure \ref{fig:intersection-density-plots}. This illustration visualizes the outputs of different models computed off-policy on a recording of the vehicle passing an intersection. Furthermore, we see the classifier model would swerve to the left if given control of the car near the intersection.

EBM models exhibit slight multimodality when passing the intersection with an area of lower energy values towards the left, corresponding to a left turn (Figure \ref{fig:intersection-density-plots} (d-f)). The standard EBM also shows an alternative hypothesis of turning to the right around 3 seconds into the recording. As reported above, however, this ability to represent alternative hypotheses in the energy landscape does not translate into improved performance.

\begin{wraptable}{r}{0.5125\linewidth}
\vspace{-0.5cm}
\caption{\textbf{Handling multimodalities.} Swerve rate shows the percentage of challenging road sections where the model slightly swerved towards the side road. There are three such road sections, with three on-policy runs per model type. Slight swerves are difficult to set a threshold for and are counted as half a swerve. A lower number is better.\\}
\label{table:multimodality-quantitative}
\centering
\begin{tabular}{lr}
\toprule
          Model                     & Swerve rate \\
\midrule
          EBM                       & 44\% \\
          EBM Temp. Smoothing       & 66\% \\
          EBM Soft Targets          & 39\% \\
          Regression                & 89\% \\
          Classification            & 44\% \\
          MDN                       & 33\% \\
\bottomrule
\end{tabular}
\vspace{-0.3cm}
\end{wraptable}

Supposedly, a model having an explicit representation of alternative paths can select the most likely and not produce intermediate behavior that is average of multiple options. At intersections, such average behavior would show as swerving towards the side road. When quantified (see Table \ref{table:multimodality-quantitative}), classical regression models were shown to produce most swerving towards side roads. This observation aligns with the unimodal nature of regression models trained with the MAE loss. In contrast, other models swerved less, which can be attributed to their richer representations.

\section{Discussion}

This project investigated if the reported benefits of using the energy-based model formulation for behavioral cloning carry over to the task of real-world road following. We hypothesized that the claimed better generalization and handling of multimodalities could be useful in this task. However, the results show no improvements in overall driving ability. We believe the chosen task has too few multimodalities to make EBMs stand out. Prior work on implicit behavior cloning\cite{florence2021implicit} used tasks where actions are distributed less normally than in road-following steering control (see Appendix Figure \ref{fig:tasks-action-distributions}). Past tasks also had at least two-dimensional action spaces, which may have more multimodalities. However, our models only predicted steering, and representing different steering hypotheses was useful only at a few intersections along the route.

We did observe some improvement in handling situations that presumably require modeling multimodal distributions, such as intersections and ignoring side roads. The only unimodal baseline, MAE-based regression, swerved towards side roads more frequently than other models. However, less frequent swerving of multimodal models did not result in fewer interventions overall. We attribute this to the low proportion of the task requiring a multimodal policy. Conversely, the unimodal loss seems to introduce an inductive bias, increasing the data efficiency in learning to handle simpler (unimodal) road situations that dominate the task.

We observed higher lateral jerk for multimodality-representing models. Our proposed modifications to the EBM training process significantly reduced the jerk of the model-predicted commands. However, this did not improve the effective whiteness of the car's front wheels. We attribute this to the car's actuators acting as a low-pass filter on the noise in the command sequence. This prevents even a large change in command whiteness from translating to a drop in front-wheel whiteness, motivating research on more powerful smoothing techniques. The temporal loss term was computed on consecutive images only 33 ms apart. However, effective whiteness seems to be caused by output variability on a slightly higher time scale. In future work, smoothing actions across a slightly longer timescale should be attempted. Soft targets proved surprisingly effective in reducing command-sequence whiteness, given that they do not directly enforce similarity across time. 

\section{Conclusion}

We tested implicit behavioral cloning with energy-based models for controlling the steering of a real self-driving car. We showed that energy-based models perform comparably to classical explicit behavioral cloning baselines in terms of safety driver interventions but have higher jerk that reduces the comfort of the drive. We show two methods for reducing the steering jerk, measured as a whiteness score. Even though these methods greatly reduce the whiteness of predicted steering angles, it does not translate into improved whiteness of real steering, as the actuator delays in a real car smooth out radical steering movements anyway.

In our experiments, the simple regression-based explicit behavioral cloning baseline was the best in terms of interventions and jerk of the drive. However, the regression approach tended to swerve towards side roads, which comes from the unimodal nature of its loss function. We show that multimodality-capable models handle the situation with side roads better and do fewer swerves but do not eliminate the problem. Based on the analysis of action distributions in our task and in prior work where EBMs outperformed explicit models, we conclude that the lateral control in the road-following task has too few multimodalities to make EBMs useful. Altogether, this shows that while energy-based models have a number of theoretical advantages, it can be challenging to bring those out in real-world scenarios, and more research is needed to make efficient use of them.
\newpage
\begin{ack}
This work was supported by the collaboration project LLTAT21278 with Bolt Technologies, and by the Estonian Research Council grant PRG1604. Mikita Balesni was funded by the Estonian Research Council grant for Ukrainian Researchers.
\end{ack}

{
\small

\bibliography{citations}

}


\newpage
\appendix

\section{Appendix}

\subsection{Pre-processing, training, and hardware}

\paragraph{Pre-processing} The frames are cropped to remove the car’s hood and everything beyond the horizon and to limit the view to 90 degrees of the front center. The resulting frames of shape 264x68x3 are then min-max normalized and fed into the model.

\paragraph{Training} When training, a mini-batch is created by sampling uniformly from all recordings. For experiments with temporal smoothing, a different sampling approach is used, where sequences of two consecutive frames are sampled instead. The sequence dimension is flattened such that a mini-batch of sequences becomes a mini-batch of frames. The target labels correspond to the steering wheel angles of the human drivers. 

The Adam \cite{adam-optimizer} optimizer is used with default hyperparameters (learning rate $1*10^{-3}$, betas $0.9$ and $0.999$) and $1 * 10^{-2}$ weight decay \cite{weight-decay}. Finally, early stopping is used on validation MAE with a patience of 10 epochs.

\paragraph{Car Hardware and Software Stack}

We perform the experiments with Lexus RX 450h fitted with a PACMod v3 drive-by-wire system. The following sensors are used: a NovAtel PwrPak7D-E2 GNSS device and a Sekonix SF3324 120-degree FOV camera. The car computer is equipped with a GeForce GTX 2080 GPU. The camera works at 30 Hz, but our end-to-end stack is slower ($\sim$12 Hz). To accommodate for the differing processing speeds all but the latest frame in the queue are dropped.

\subsection{Validation loss and whiteness with different data amounts}

\begin{figure}[h]
    \centering
    \includegraphics[width=0.49\linewidth]{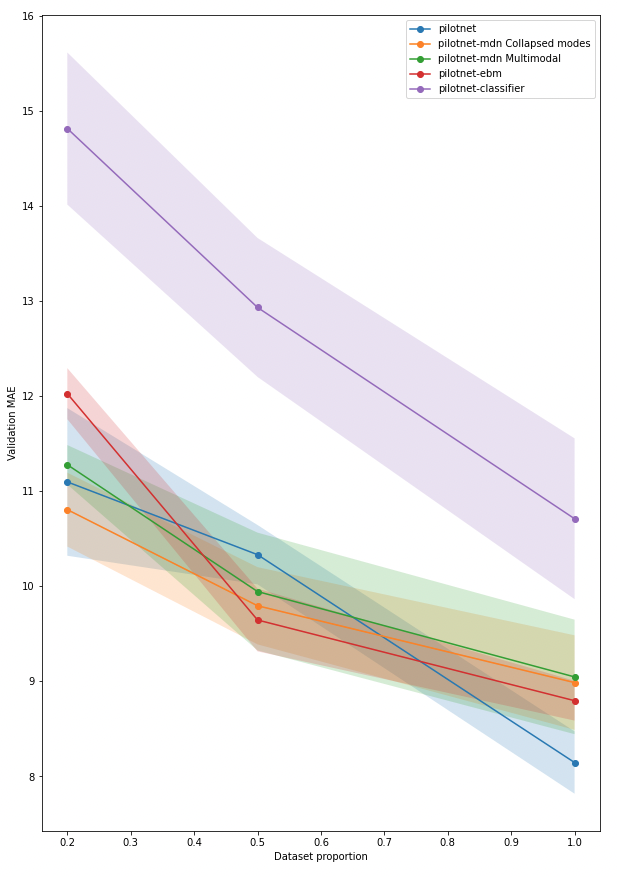}
    \includegraphics[width=0.49\linewidth]{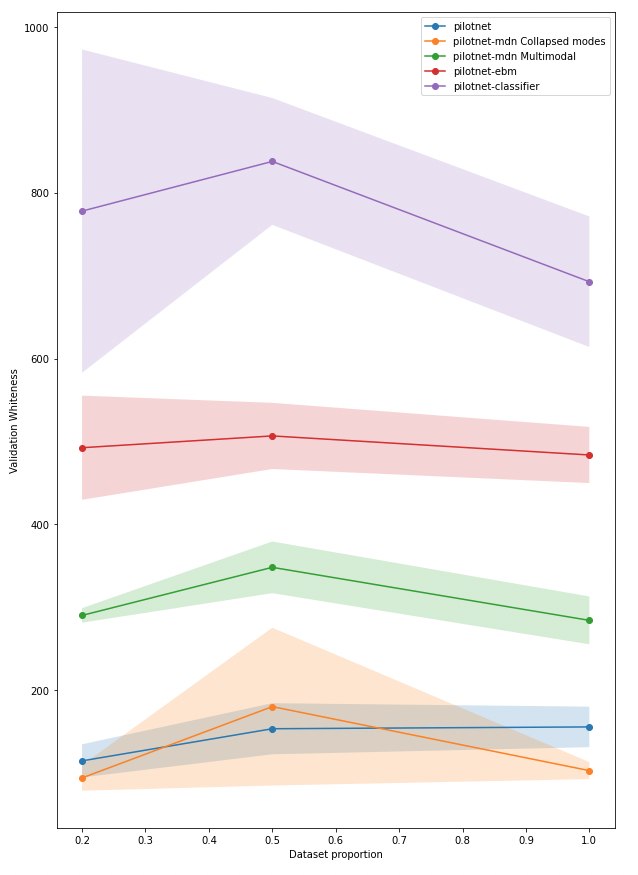}
    \caption{\textbf{Left:} Varying the dataset size intuitively changes the predictive accuracy of all model formulations. \textbf{Right:} In contrast, whiteness does not seem to be influenced by the amount of data when the amount was varied by 5 times.}
    \label{fig:mae-whiteness-scaling}
\end{figure}

\subsection{How do our EBM simplifications affect modeling performance?}

\begin{figure}[H]
    \centering
    \includegraphics[width=0.49\linewidth]{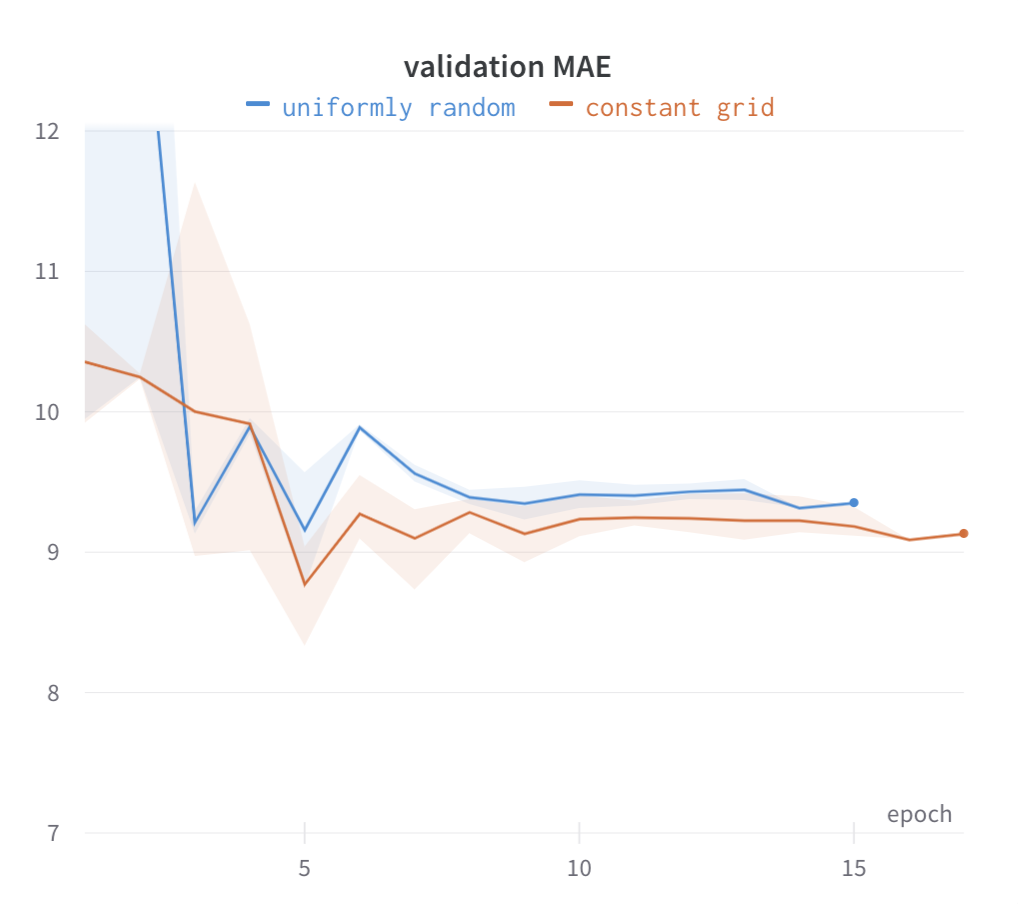}
    \includegraphics[width=0.49\linewidth]{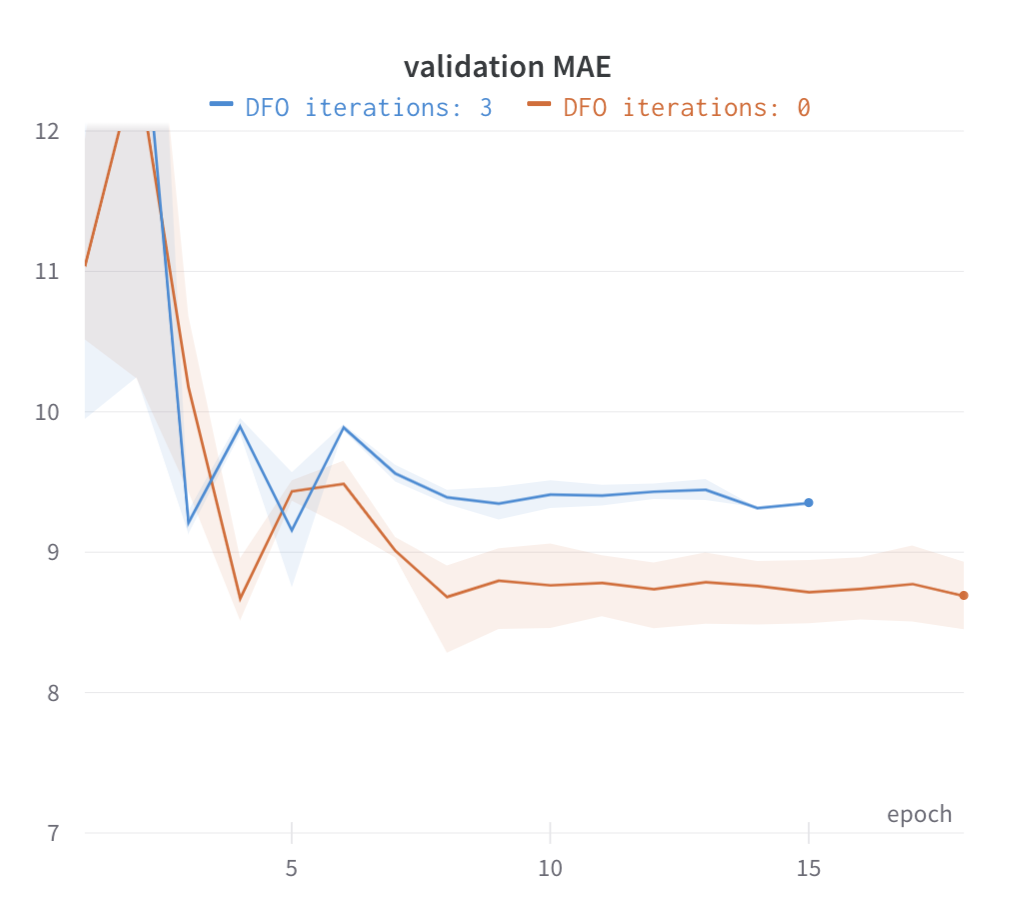}
    \caption{\textbf{Left:} Using a constant grid of actions results in at least as good MAE as the more common uniformly random sample on each decision. \textbf{Right:} Derivative-free optimization does not improve performance beyond a one-shot argmin; to our surprise, it even hurts validation MAE.}
    \label{fig:ebm-variants-performance}
\end{figure}

\subsection{Is road following just too unimodal?}

\begin{figure}[h]
    \centering
    \begin{subfigure}[b]{0.4\textwidth}
        \centering
        \includegraphics[width=\textwidth]{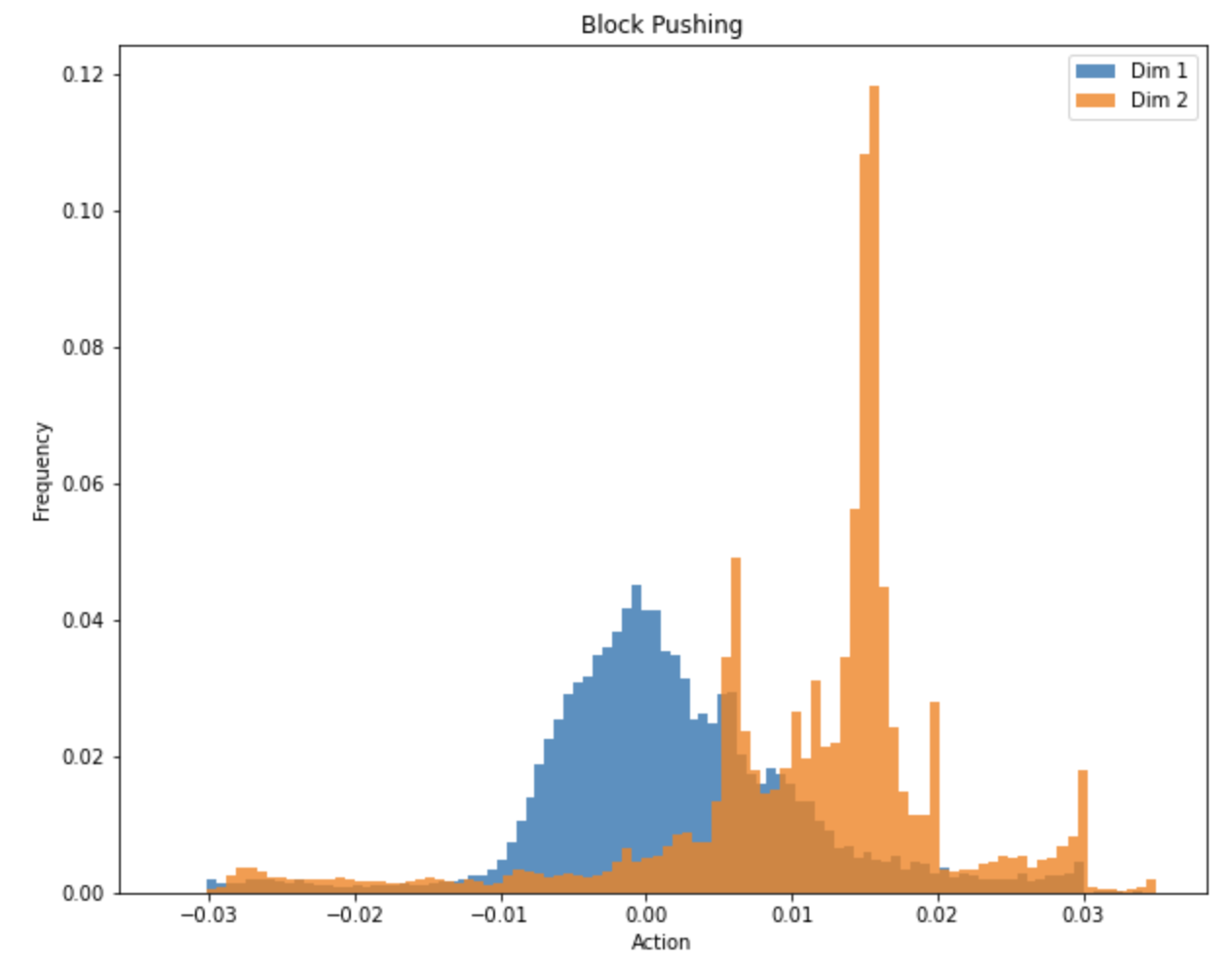}
        \caption{}
        \label{fig:tasks-action-distribution-a}
    \end{subfigure}
    \begin{subfigure}[b]{0.4\textwidth}
        \centering
        \includegraphics[width=\textwidth]{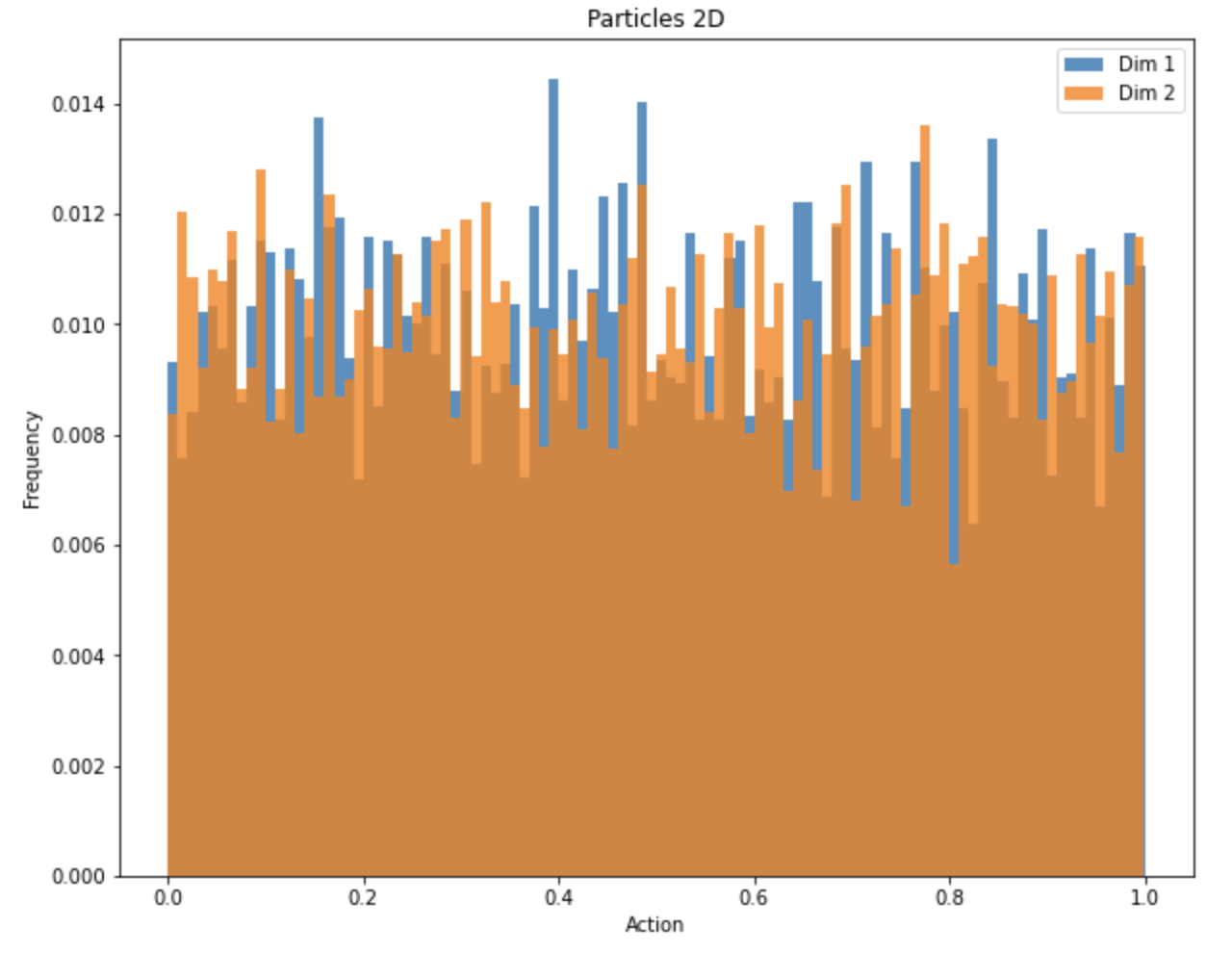}
        \caption{}
        \label{fig:tasks-action-distribution-b}
    \end{subfigure}
    \begin{subfigure}[b]{0.4\textwidth}
        \centering
        \includegraphics[width=\textwidth]{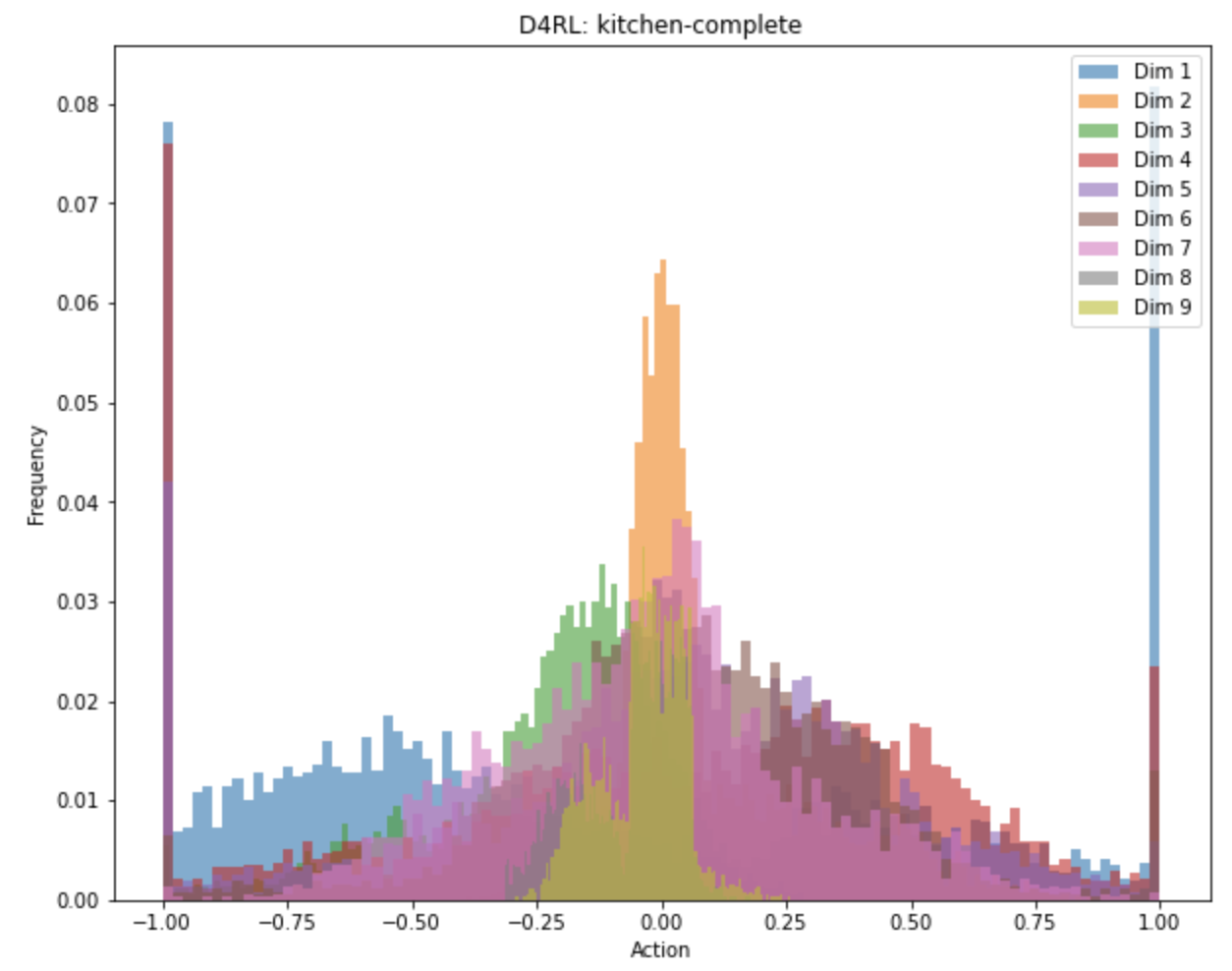}
        \caption{}
        \label{fig:tasks-action-distribution-c}
    \end{subfigure}
    \begin{subfigure}[b]{0.4\textwidth}
        \centering
        \includegraphics[width=\textwidth]{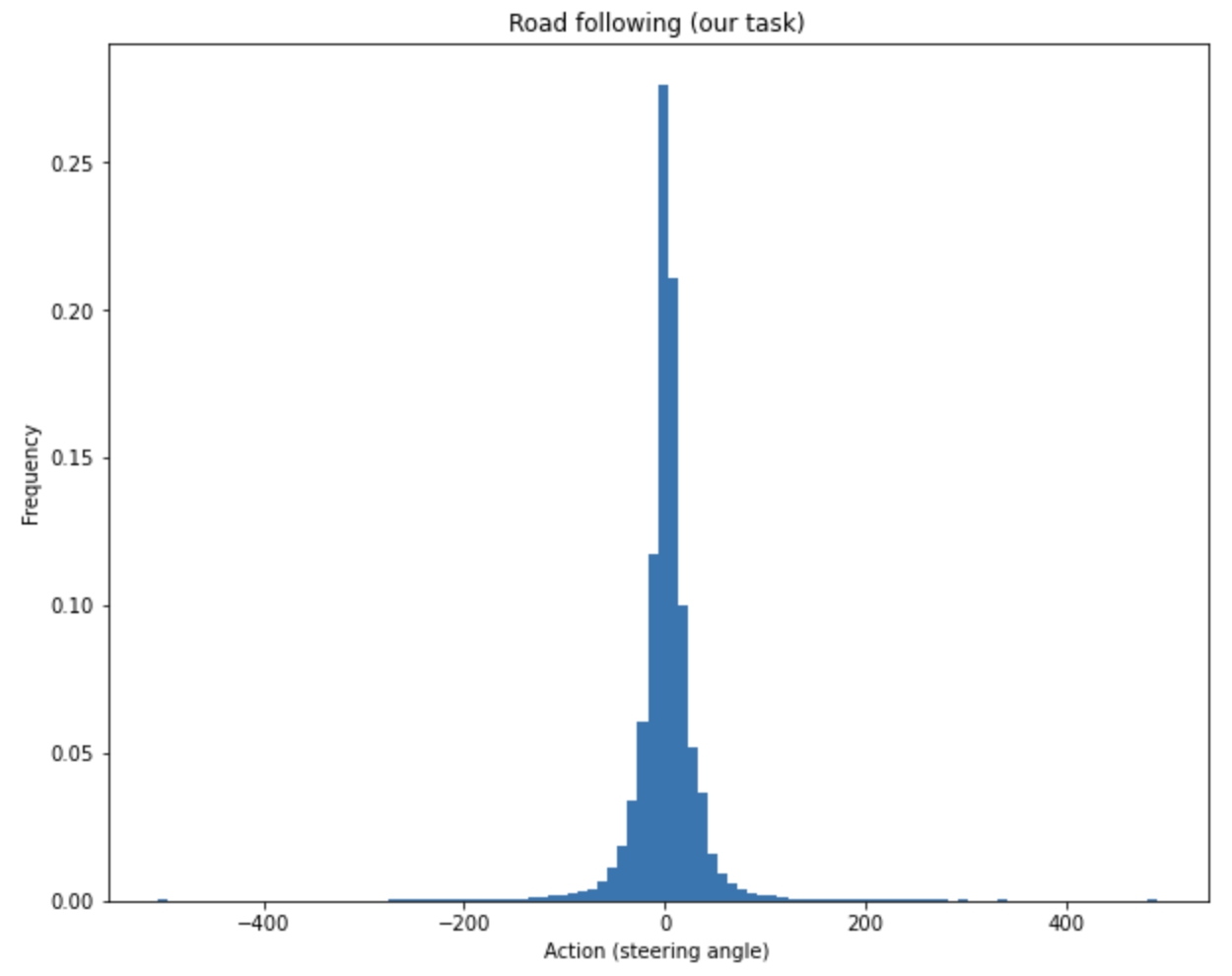}
        \caption{}
        \label{fig:tasks-action-distribution-d}
    \end{subfigure}
    \caption{Action distributions from three randomly-picked example tasks from prior work on IBC\cite{florence2021implicit} where EBMs outperformed explicit models \textbf{(a-c)}. Action distribution for our road-following task \textbf{(d)} looks much more gaussian, which is a hint for a lower number of possible multimodalities.}
    \label{fig:tasks-action-distributions}
\end{figure}

\newpage
\subsection{VISTA agreement with real-world results}

In the main text, we report VISTA's agreement with the results in the main experiments. In Table \ref{table:vista-correlations}, we report the results for all models we tested throughout the project. When run on the same track, VISTA seems to have a very high agreement with real-world results.

\begin{table}[h]
\vspace{-0.3cm}
\caption{Per-model mean metrics correlations for VISTA vs reality (n=17 models).}
\label{table:vista-correlations}
\centering
\begin{tabular}{lll}
\toprule
        Measure     & Interventions & $W_{cmd}$   \\
        \midrule 
        Pearson     & 83\%                  & 89\% \\
        Spearman    & 84\%                  & 86\% \\
\bottomrule
\vspace{-0.3cm}
\end{tabular}
\end{table}

\newpage
\subsection{Notepad}

\begin{table}[h]
\caption{Generalization results, with three real-world and three virtual driving sessions per model.}
\label{generalization-results-table}
\centering
\begin{tabular}{lllll}
\toprule
          &  \multicolumn{2}{c}{Real world} & \multicolumn{2}{c}{VISTA} \\
          \cmidrule(r){2-3} \cmidrule(r){4-5}
          Model &  Interventions & Whiteness & Crashes & Whiteness \\
\midrule
          \multirow{3}{*}{EBM} & 4  & 176.93°/s & 2    & 114.33°/s \\
           &                      1 & 96.94°/s  & 1    & 121.57°/s \\
           &                      2 & 223.59°/s & 2    & 121.67°/s \\
          \cmidrule(r){2-3} \cmidrule(r){4-5}
          \qquad\qquad\qquad mean: &  2.33 & 165.82°/s & 1.67 & 119.19°/s \\

          \midrule
          \multirow{3}{*}{EBM Temp. Smoothing} &   5  & 119.39°/s & 3 & 58.70°/s \\
           &                                       2  & 137.22°/s & 2 & 60.37°/s \\
           &                                       3  &  77.28°/s & 2 & 48.86°/s  \\
           \cmidrule(r){2-3} \cmidrule(r){4-5}
           \qquad\qquad\qquad mean:                &   3.33  & 111.30°/s & 2.33 & 55.98°/s \\

           \midrule
           \multirow{3}{*}{EBM Soft Targets}     &  5 & 56.33°/s & 3 & 85.72°/s \\
           &                                        5 & 57.15°/s & 3 & 74.97°/s \\
           &                                        4 & 56.86°/s & 3 & 81.87°/s \\
           \cmidrule(r){2-3} \cmidrule(r){4-5}
           \qquad\qquad\qquad mean:                  &  4.66 & 56.78°/s & 3 & 80.85°/s \\

           \midrule
           \multirow{3}{*}{Regression (MAE)} &      2 & 37.84°/s & 0 & 24.39°/s \\
           &                                        2 & 75.34°/s & 0 & 24.75°/s \\
           &                                        1 & 33.10°/s & 0 & 24.25°/s \\
           \cmidrule(r){2-3} \cmidrule(r){4-5}
           \qquad\qquad\qquad mean:                  &  1.66 & 48.76°/s & 0 & 24.47°/s \\

           \midrule
           \multirow{3}{*}{Classification} &        1 & 182.39°/s & 1 & 123.69°/s \\
           &                                        7 & 287.14°/s & 1 & 105.13°/s \\
           &                                        1 & 162.27°/s & 1 & 104.31°/s \\
           \cmidrule(r){2-3} \cmidrule(r){4-5}
           \qquad\qquad\qquad mean:                   & 3.00 & 210.60°/s & 1 & 111.04°/s \\
         
           \midrule
           \multirow{3}{*}{MDN} &                   1 & 33.62°/s & 3 & 37.22°/s \\
           &                                        5 & 35.46°/s & 3 & 35.74°/s \\
           &                                        5 & 37.39°/s & 3 & 35.84°/s \\
           \cmidrule(r){2-3} \cmidrule(r){4-5}
           \qquad\qquad\qquad mean:                   & 3.66 & 35.49°/s & 3 & 36.27°/s \\
\bottomrule
\end{tabular}
\end{table}

\end{document}